\documentclass{article}
\usepackage[preprint,nonatbib]{neurips_2026}
\usepackage{amsthm}
\usepackage{bbm}
\newtheorem{theorem}{Theorem}
\newtheorem{lemma}{Lemma}
\newtheorem{proposition}{Proposition}
\newtheorem{corollary}{Corollary}

\newtheorem{remark}{Remark}

\newtheorem{assumption}{Assumption}

\usepackage{enumitem}
\usepackage[utf8]{inputenc} % allow utf-8 input
\usepackage[T1]{fontenc}    % use 8-bit T1 fonts
\usepackage{url}            % simple URL typesetting
\usepackage{booktabs}       % professional-quality tables
\usepackage{amsfonts}       % blackboard math symbols
\usepackage{nicefrac}       % compact symbols for 1/2, etc.
\usepackage{microtype}      % microtypography
\usepackage{xcolor}

\usepackage[numbers]{natbib}
\usepackage[hidelinks]{hyperref}
\usepackage{siunitx}
\usepackage{bm}
\usepackage{comment}
\usepackage{wrapfig}
\usepackage{graphicx}
\usepackage{amsmath}
\usepackage{amssymb}
\usepackage{xspace}
\usepackage{algorithm}
\usepackage{algpseudocode}
\algrenewcommand\algorithmiccomment[1]{\hfill$\triangleright$ #1}

\usepackage{tabularx}
\usepackage{multirow}
\usepackage{subcaption}
\usepackage{booktabs}
\usepackage{xspace}
\usepackage[capitalise]{cleveref}

\crefname{theorem}{Theorem}{Theorems}
\crefname{lemma}{Lemma}{Lemmas}
\crefname{proposition}{Proposition}{Propositions}
\crefname{corollary}{Corollary}{Corollaries}
\crefname{example}{Example}{Examples}
\crefname{remark}{Remark}{Remarks}
\crefname{definition}{Definition}{Definitions}
\crefname{assumption}{Assumption}{Assumptions}

\newcommand{\E}{\mathbb{E}}
\newcommand{\R}{\mathcal{R}}

\newcommand{\Kls}{\mathsf{K}_{\lambda,s}}

\newcommand{\ri}{\operatorname{ri}}
\newcommand{\lin}{\operatorname{span}}

\newcommand{\tr}{\operatorname{tr}}

\DeclareMathOperator*{\argmin}{arg\;min}
\usepackage{makecell}
\usepackage{bbm}
\usepackage{wrapfig}
\usepackage{graphicx}
\usepackage{ifthen}
\usepackage{xspace}
\usepackage{upgreek}
\usepackage{pifont}% http://ctan.org/pkg/pifont
\usepackage{colortbl}
\usepackage{multirow}
\usepackage{dsfont}
\usepackage{float}

\usepackage{tikz}
\usetikzlibrary{patterns}
\usepackage{pgfplots}
\pgfplotsset{compat=newest}
\pgfplotsset{every axis legend/.append style={cells={anchor=west}}}
\usetikzlibrary{arrows}
\tikzset{>=stealth'}    % colors

\title{The Curvature of Regret in Contextual Linear Optimization}

\author{%
  Konstantinos Ziliaskopoulos\\
  Auburn University\\
  Auburn, AL, USA\\
  \texttt{kzz0034@auburn.edu}
  \And
  Alexander Vinel\\
  Auburn University\\
  Auburn, AL, USA\\
  \texttt{alexander.vinel@auburn.edu}
  \And
  Alice E. Smith\\
  University of Alabama\\
  Tuscaloosa, AL, USA\\
  \texttt{alice.smith@ua.edu}
}
\begin{document}

\maketitle

\begin{abstract}
Decision-focused learning for linear optimization is complicated by the discontinuity of the optimizer, where small cost errors may leave the decision unchanged or move it to a different vertex. We show that this non-smooth pointwise behavior becomes locally quadratic after averaging over the data distribution, and we derive the curvature in closed form, specifically, a matrix-valued measure supported on the walls of the normal fan. This measure depends only on the feasible set, with the data distribution entering only as a weight. We then offer a tractable approximation for this curvature, computable with just one projection to the feasible set.  We prove that the approximation weakly converges to the true population curvature. We offer one application of our findings, a decision-aware scenario generation method for expected-cost linear optimization. Our experiments test the quadratic and weak convergence laws and show a 30.8\% regret improvement over uniform allocation on battery arbitrage.
\end{abstract}

\vspace{-0.5em}
\section{Problem Setup and Main Results}
\label{sec:introduction}

In standard Predict-then-Optimize systems, we first forecast unknown objective coefficients and then solve a linear program (LP) \citep{elmachtoub2022smart,mandi2024dfl,sadana2025survey,mivsic2020data}. Let $S\subseteq\mathbb R^d$ be a compact convex polytope representing the feasible set. Without loss of generality, we discard cost directions orthogonal to \(\operatorname{span}(S-S)\), since they are constant over \(S\) and do not affect the optimizer. Then define
\(
w^\star(c)\in\arg\min_{w\in S}c^\top w,
\) and
\(
z_S(c)=\min_{w\in S}c^\top w.
\)
For data $(X,C)$ and a cost predictor $f$, let
$R(f)=\mathbb E[C^\top w^\star(f(X))-z_S(C)]$ represent the LP regret.
The relevant cost vector for regret is the conditional mean $Q=\mathbb E[C\mid X]$, not $\mathbb E[C]$. If $R_X^\star = \inf_{g:g(X)\in S} \mathbb{E}[C^Tg(X)-z_s(C)]$ is the best regret attainable by any decision determined by $X$, then (see \cref{lem:conditional-mean} in Appendix A for details)
\[
R(f)-R_X^\star
=
\mathbb E\!\left[Q^\top w^\star(f(X))-z_S(Q)\right].
\]

Thus, \(R_X^\star\) is the best regret achievable when the decision can depend only on X. The remaining excess regret depends on whether the decision selected using $f(X)$ is suboptimal under $Q=\mathbb E[C\mid X]$.

In this paper, we focus on polyhedral feasible sets $S$. For each vertex $v$ of $S$, define its optimizer cell \(\mathcal C_v=\left\{q\in\mathbb R^d:q^\top v=z_S(q)\right\}\), the region in cost space where $q$ induces the decision $v$. Two full-dimensional cells share a wall if and only if their corresponding vertices are adjacent in $S$. For adjacent vertices $v_i$ and $v_j$, define the wall \(W_{ij} = \left\{ q: q^\top v_i=q^\top v_j=z_S(q) \right\}\) and jump vector \( d_{ij}=v_i-v_j\), where $W_{ij}\subseteq d_{ij}^{\perp}$. Let $\mathcal W(S)$ denote the collection of walls. For $W\in\mathcal W(S)$ let $d_W$ be either orientation of its jump vector. Under local density and smoothness assumptions, suppose that $Q$ has density $p$ and that $f_\varepsilon(X)=Q+\varepsilon h(Q)$, where $\varepsilon h(Q)$ is a small prediction error. Then, in this paper, we show that 
\vspace{-0.5em}
\begin{equation}
\label{eq:wp-definition}
R(f_\varepsilon)-R_X^\star = 
\frac{\varepsilon^2}{2}\,\kappa_p(h)+o(\varepsilon^2), \qquad
\kappa_p(h) = 
\sum_{W\in\mathcal W(S)}
\int_W
\frac{p(q)}{\|d_W\|}
\bigl(d_W^\top h(q)\bigr)^2
\,\mathrm d\mathcal H^{d-1}(q). \footnotemark
\end{equation}
\footnotetext{\(\mathcal H^{d-1}\) is the Hausdorff measure, the $(d-1)$-dimensional measure of the wall, length in 2D, area in 3D, etc.} The intuition behind this local quadratic behavior is that a small perturbation must both reach a wall and cross it. An $O(\varepsilon)$ perturbation can change the optimal decision only when the cost vector lies within $O(\varepsilon)$ of a wall, as illustrated in \cref{fig:vignette}. Each wall crossing incurs $O(\varepsilon)$ regret, so the expected regret is $O(\varepsilon)\times O(\varepsilon)=O(\varepsilon^2)$. We therefore show that curvature is concentrated on the walls of the optimizer fan. The integrand in \cref{eq:wp-definition} can also be viewed as a decomposition of prediction error into the three quantities that determine decision relevance: (i) $p(q)$, the density of the conditional cost at the wall, (ii) $\|d_W\|$, the severity of the decision switch, and (iii) $\left(\nicefrac{d_W^T h(q)}{\|d_W\|}\right)^2$, the component of the error that is orthogonal to the wall. Since the curvature in \cref{eq:wp-definition} is concentrated on the walls, a set of zero volume, it does not appear as an ordinary Hessian at individual cost vectors, and
\begin{wrapfigure}{r}{0.45\textwidth}
    \vspace{-12pt}
    \centering
    \scalebox{1}[0.85]{%
        \begin{tikzpicture}[x=2.05cm,y=2.05cm,line join=round,line cap=round,font=\footnotesize]
 
  %------------------------------------------------------------------
  % optimizer cells of the normal fan (clipped to [-1.4,1.4]^2)
  %------------------------------------------------------------------
  \fill[black!3]  (0,0) -- (1.4,0)   -- (1.4,1.4)   -- (0,1.4)  -- cycle; % w* = (0,0)
  \fill[black!8]  (0,0) -- (0,1.4)   -- (-1.4,1.4)  -- (-1.4,-1.4) -- cycle; % w* = (1,0)
  \fill[black!12] (0,0) -- (-1.4,-1.4) -- (1.4,-1.4) -- (1.4,0)  -- cycle; % w* = (0,1)
 
  %------------------------------------------------------------------
  % O(eps) buffer around the wall q1 = q2  (half-width 0.1 along (1,-1)/sqrt2)
  %------------------------------------------------------------------
  \fill[black!22] (-0.139,-0.281) -- (-1.217,-1.359) --
                  (-1.359,-1.217) -- (-0.281,-0.139) -- cycle;
 
  %------------------------------------------------------------------
  % coordinate axes (thin; the positive semi-axes are themselves walls)
  %------------------------------------------------------------------
  \draw[thin,black!55] (-1.4,0) -- (0,0);
  \draw[thin,black!55] (0,-1.4) -- (0,0);
  \node[anchor=west]  at (1.44,0) {$q_1$};
  \node[anchor=south] at (0,1.44) {$q_2$};
 
  %------------------------------------------------------------------
  % decision boundaries: the walls of the fan
  %------------------------------------------------------------------
  \draw[line width=1.4pt] (0,0) -- (1.4,0);
  \draw[line width=1.4pt] (0,0) -- (0,1.4);
  \draw[line width=1.4pt] (0,0) -- (-1.4,-1.4);
 
  %------------------------------------------------------------------
  % feasible set S = conv{(0,0),(1,0),(0,1)}, overlaid
  %------------------------------------------------------------------
  \fill[pattern=north east lines,pattern color=black!40] (0,0) -- (1,0) -- (0,1) -- cycle;
  \draw[line width=0.8pt] (0,0) -- (1,0) -- (0,1) -- cycle;
  \foreach \p in {(0,0),(1,0),(0,1)} \fill \p circle (1.7pt);
  \node at (0.42,0.42) {$S$};
  \node[anchor=south west,inner sep=1.5pt] at (0.05,0.03) {$(0,0)$};
  \node[anchor=north] at (1,-0.06) {$(1,0)$};
  \node[anchor=east]  at (-0.05,1) {$(0,1)$};
 
  %------------------------------------------------------------------
  % buffer outline, width marker and wall annotation
  %------------------------------------------------------------------
  \draw[densely dotted,black!70] (-0.139,-0.281) -- (-1.217,-1.359);
  \draw[densely dotted,black!70] (-0.281,-0.139) -- (-1.359,-1.217);
  \draw[<->,>=stealth,thin] (-0.881,-1.023) -- (-1.023,-0.881);
  \draw[thin] (-0.881,-1.023) -- (-0.845,-1.115);
  \node[anchor=west,inner sep=1.5pt] at (-0.835,-1.13) {$O(\varepsilon)$};
  \node[anchor=east] at (-0.62,-0.32) {$q_1=q_2$};
 
  %------------------------------------------------------------------
  % which vertex each cell selects
  %------------------------------------------------------------------
  \node at (0.82,1.05)  {$w^\star=(0,0)$};
  \node at (-0.78,0.75) {$w^\star=(1,0)$};
  \node at (0.62,-0.80) {$w^\star=(0,1)$};
 
\end{tikzpicture}
    }
    \caption{The normal fan of a triangle feasible set. In the shaded $O(\varepsilon)$ neighborhood of the wall $q_1=q_2$ we show that an $O(\varepsilon)$ perturbation can change the decision only for cost vectors inside it, and each crossing incurs $O(\varepsilon)$ regret, which is the source of local quadratic behavior.
}
    \label{fig:vignette}
\vspace{-36pt}
\end{wrapfigure}
is instead represented by the matrix-valued measure,
\begin{equation}
\label{eq:K0-definition}
K_0
=
\sum_{W\in\mathcal W(S)}
\frac{d_Wd_W^\top}{\|d_W\|}
\,\mathcal H^{d-1}\!\restriction_W.
\end{equation}
We can also write $\kappa_p = \int_{\mathbb{R}^d}p(q)h(q)^\top dK_0 h(q)$. Because $K_0$ is singular and supported on the walls, it is difficult to estimate directly. We therefore construct a smooth approximation using a quadratically regularized decision rule. Choose a reference point \(s\) in the relative interior of \(S\) and a regularization parameter \(\lambda > 0\). We define,
\begin{equation}
\label{eq:w_lambda-def}
w_{\lambda,s}(q)
=
\arg\min_{w\in S}
\left\{q^\top w+\frac{\lambda}{2}\|w-s\|^2\right\}
,
\end{equation}
which also corresponds to the projection to the feasible set $w_{\lambda,s}(q) = \Pi_S\!\left(s-\frac q\lambda\right)$.
Its \emph{regularized curvature matrix} is
\begin{equation}
\label{eq:K-curvature-def}
\mathsf K_{\lambda,s}(q) = -D_qw_{\lambda,s}(q) =
\frac1\lambda J\Pi_S\!\left(s-\frac q\lambda\right),
\end{equation}

wherever the derivative (Jacobian) $D_q w_{\lambda, s}(q)$ exists. We show that as $\lambda \to 0$, $\mathsf K_{\lambda,s}\rightharpoonup K_0$, and therefore \(\mathbb E\!\left[h(Q)^\top \mathsf K_{\lambda,s}(Q)h(Q)\right]=\kappa_p(h)+O(\lambda).\)
We also show that our approximation is computationally efficient to estimate. We can identify the active face with one projection, after which the curvature matrix \(\mathsf K_{\lambda,s}(q)\) can be obtained by scaling the orthogonal projector onto the directions parallel to that face by $1/\lambda$.

Prior work on contextual linear optimization has characterized generalization and fast regret rates using margin conditions that control predictions or costs near decision boundaries \citep{tsybakov2004optimal,hu2022fast,elbalghiti2023generalization}. We build on this perspective by identifying the exact local second-order behavior of population regret. Decision-focused learning has also developed local quadratic and metric losses that emphasize decision-relevant prediction errors \citep{shah2022lodl,bansal2023taskmet}, and quadratic regularization and projection have been used to construct differentiable prediction maps and losses \citep{wilder2019melding,martins2016softmax,niculae2018sparsemap,blondel2020learning}, including recent work that computes regret gradients as a projection of prediction error \citep{lee2026pear}. For LPs these constructions coincide with \(\mathsf K_{\lambda,s}\). We identify the object these methods approximate at the \(\lambda \downarrow 0\) limit in \cref{thm:weak} as the measure \(K_0\).

The contributions of this research can be summarized as follows. %\textbf{Population quadratic behavior.} 
First, we show that although pointwise LP regret is discontinuous, population regret is locally quadratic, and derive the closed-form curvature concentrated on the optimizer walls. %\textbf{Curvature representation and tractable approximation.} 
We then represent this curvature with $K_0$ and show that the regularized curvature $\mathsf K_{\lambda,s}$ converges weakly to $K_0$. Furthermore, we show how to derive this surrogate using just a single projection (see \cref{prop:oneproj} in the Appendix). 
%\textbf{Decision-aware scenario allocation.} 
Finally, we use the curvature to derive a local bias-variance decomposition for generated scenarios for stochastic programs and a curvature-weighted allocation rule.

We now formalize these observations and state our main theoretical results. We assume standard regularity conditions for $h(q)$ and $p(q)$ and bounded perturbations $h(q)$. We provide further details in \cref{ass:local} in Appendix A, along with the corresponding proofs.

\begin{theorem}[The quadratic law and the closed form of
\(\kappa_p\)]\label{thm:quadratic-law}
Suppose Assumption~\ref{ass:local} holds and let
\(
f_\varepsilon(X):=Q+\varepsilon h(Q)\) where $\varepsilon \downarrow 0$.
Then
\[
\R(f_\varepsilon)-\R_X^\star
=
\frac{\varepsilon^2}{2}\kappa_p(h)
+
o(\varepsilon^2).
\]
Equivalently,
\[
\lim_{\varepsilon\downarrow0}
\frac{2\bigl(\R(f_\varepsilon)-\R_X^\star\bigr)}
{\varepsilon^2}
=
\kappa_p(h).
\]
Hence from the definition of $\kappa_p$ in \cref{eq:wp-definition},  the population curvature is supported on the codimension-one walls of the normal fan.\footnote{Here codimension-one means one dimension lower than the ambient space $\mathbb{R}^d$.}
\end{theorem}

Because this measure is supported on a measure-zero set, we approximate it by the regularized curvature $K_{\lambda,s}$ in \cref{eq:K-curvature-def}. The point $x = s-q/\lambda$ projects onto some face $F$ of $S$, and the Jacobian of a Euclidean projection onto a polytope is the orthogonal projector onto the directions along $F$. A cost vector within $O(\lambda)$ of a wall $F_E$ projects onto the edge $E$, where that projector is rank one along $d_E$. So on a tube of width $O(\lambda)$ around each wall, $K_{\lambda,s}$ is $1/\lambda$ times a rank-one matrix. With width $O(\lambda)$ times height $1/\lambda$ is $O(1)$, and $K_{\lambda,s}$ is a smoothed surrogate of the wall density.

\begin{theorem}[Regularized curvature approximates the population curvature]
\label{thm:weak}

Under \cref{ass:local}, with $\kappa_p(h)$ defined in \cref{eq:wp-definition}, as $\lambda \downarrow 0$, $\mathsf K_{\lambda,s} \rightharpoonup K_0$. Consequently,
\begin{equation*}
  \mathbb{E}\!\left[
    h(Q)^{\top}\mathsf K_{\lambda,s}(Q)h(Q)
  \right]
  =
  \kappa_{p}(h)+O(\lambda).
\end{equation*}
Thus, the regularized curvature $\mathsf K_{\lambda,s}$ approximates the
population curvature, with an approximation error that
decreases proportionally to $\lambda$. Combining this result with \cref{thm:quadratic-law} gives
\begin{equation*}
  R(f_{\varepsilon})-R^{\star}_{X}
  =
  \frac{\varepsilon^{2}}{2}
  \mathbb{E}\!\left[
    h(Q)^{\top}K_{\lambda,s}(Q)h(Q)
  \right]
  +o(\varepsilon^{2})
  +O(\varepsilon^{2}\lambda).
\end{equation*}
Additionally, at the limit $\lambda \rightarrow 0$, only edges contribute to the population regret, with higher dimensional faces and vertices contributing nothing.
\end{theorem}

Here, we present an application of our curvature surrogate in an allocation rule for scenario generation budgets from generative models in expected-cost LPs \citep{dupavcova2003scenario,romisch2003stability, royset2013optimal,gao2017new,shi2022dynamic}. Recent generative DFL methods use conditional generators to provide decision scenarios \citep{wang2026gendfl,zhao2026diffusion}. For an expected-cost LP driven by the sample mean of $M$ generated scenarios, the same curvature separates local regret into wall-weighted generator bias term and a $1/M$ Monte Carlo covariance term, giving us a decision-aware allocation rule that assigns more scenarios where generated dispersion aligns with the curvature on the LP walls. Unlike classical scenario allocation within a single stochastic program, we consider a fixed generation budget shared across a collection of contextual decision problems. This setting can occur when conditional scenario generation is computationally costly and the value of additional samples varies substantially across contexts.

\begin{corollary}[Bias--variance split of local regret]
\label{cor:split}
For each generator $k$, conditionally on $X$, let $\widetilde{C}_{k,1},\dots,\widetilde{C}_{k,M_k}$ be i.i.d.\ draws from
a generator $k$ with mean $\mu_k(X)$ and covariance $\Sigma_k(X)$, let $\bar{C}_{k}$ be their
average, and let $b_k(X):=\mu_k(X)-Q(X)$ be the generator bias. Let
$\eta_{k}^{2}=\|b_k\|_{L^2(P_X)}^2+\nicefrac{1}{M_k}$ be the joint error term. Also assume that the normalized perturbations satisfy uniform regularity conditions (see \cref{rem:random} in Appendix) and that \(\sup_k \mathbb E[\tr(\Sigma_k(X))]<\infty.\) Consider a sequence of generators and scenario counts for which $\eta_{k}\to 0$, then,
\begin{equation*}
  R(\bar{C}_{k})-R^{\star}_{X}
  =\underbrace{\tfrac{1}{2}\,\mathbb{E}\big[b_k^{\top}\mathsf K_{\lambda,s}(Q)\,b_k\big]}_{\text{decision-relevant bias}}
  +\underbrace{\tfrac{1}{2M_k}\,\mathbb{E}\big[\operatorname{tr}\!\big(\mathsf K_{\lambda,s}(Q)\Sigma_k\big)\big]}_{\text{decision-relevant Monte Carlo noise}}
  +o(\eta_{k}^{2})+O(\lambda\eta_{k}^{2}).
\end{equation*}
\end{corollary}

Only the second term in \cref{cor:split} depends on the number of scenarios, weighting the component of the generator's conditional variance in directions normal to the decision boundary. More scenarios are allocated for generated variance that is decision-relevant, according to the curvature. We provide the construction of the generation rule in \cref{prop:curv-gen-rule} and the implementation used in our experiments in \cref{alg:curvature-allocation} in the Appendix, where one projection per context determines \(\mathsf K_{\lambda,s}\) and scenarios are allocated proportionally to \(\sqrt{\operatorname{tr}(K_{\lambda,s}\Sigma)}\).

\vspace{-0.5em}
\section{Experiments \& Conclusion}
\label{sec:results}

\begin{figure}[hb]
    \centering
    \begin{subfigure}{0.48\linewidth}
        \centering
        \includegraphics[width=\linewidth]{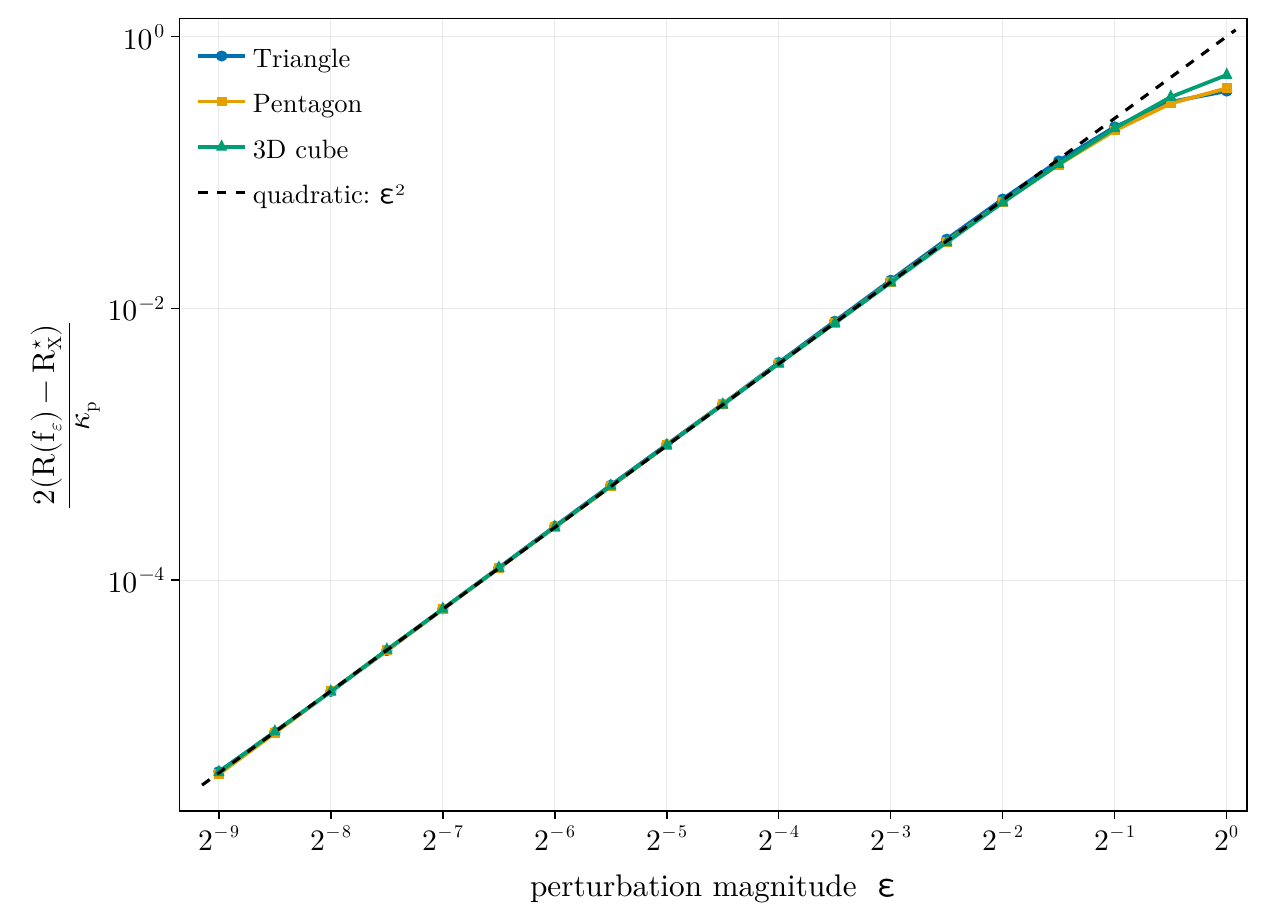}
        \caption{Local quadratic rate of population regret across three feasible set geometries.}
        \label{fig:quadratic-rate-a}
    \end{subfigure}
    \hfill
    \begin{subfigure}{0.48\linewidth}
        \centering
        \includegraphics[width=\linewidth]{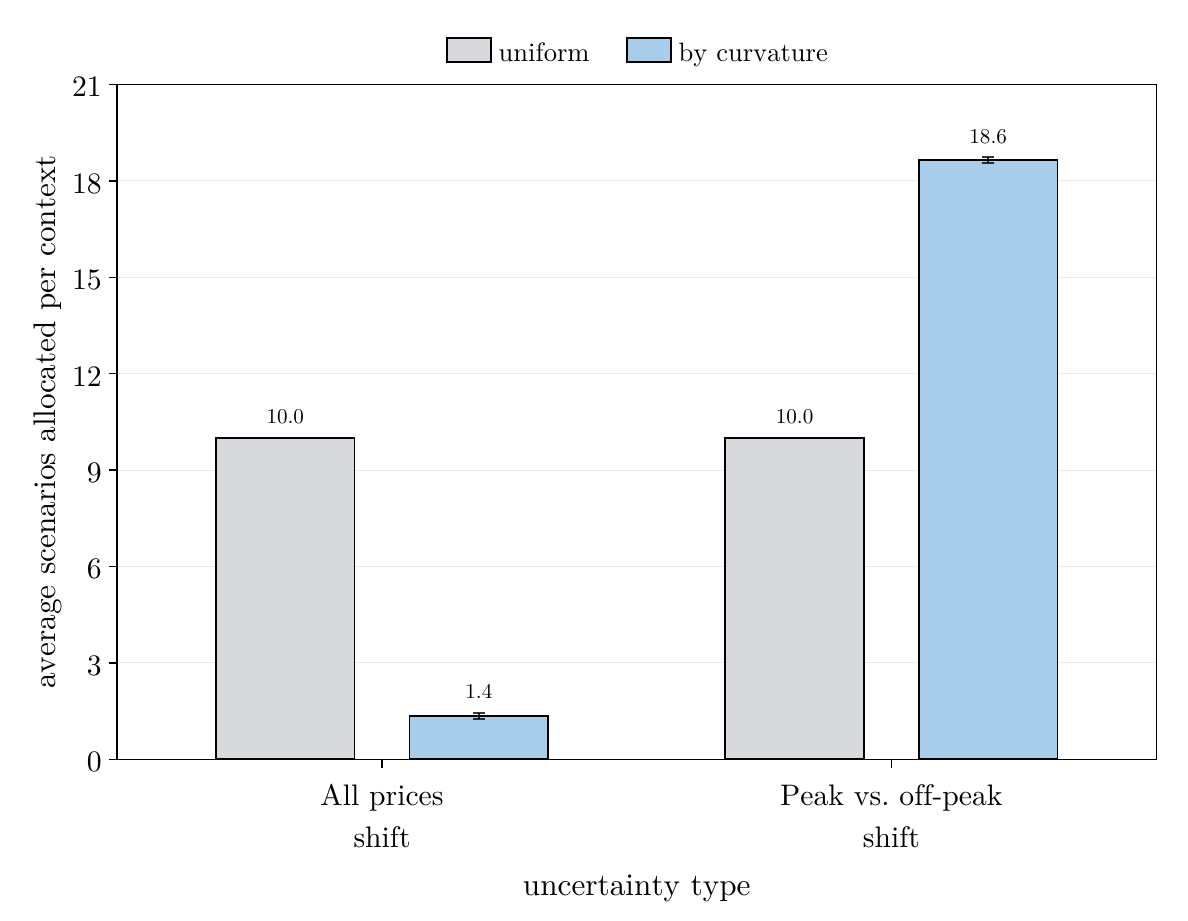}
        \caption{Comparison of uniform vs. curvature weighted generation across decision relevant/irrelevant variance.}
        \label{fig:uncertainty-alloc-b}
    \end{subfigure}
    \caption{We demonstrate the local quadratic rate of population regret and compare generation rules, showing our curvature weighted rule correctly allocates budget to relevant generative variance.}
    \label{fig:results-fig}
    \vspace{-0.5em}
\end{figure}

In \cref{fig:results-fig} we show results of two experiments: empirical validation of the local quadratic rate and performance of the curvature weighted generation rule. To test our local quadratic law of \cref{thm:quadratic-law} empirically, we sample cost vectors from smooth distributions, perturb them by a controlled amount $\varepsilon$, solve the LP, and measure the resulting excess regret for three different feasible set geometries: a 2D triangle, a 2D pentagon, and a 3D cube. We divide each empirical sampled regret with the curvature coefficient $\kappa_p$, which we compute analytically according to \cref{eq:wp-definition}, and compare it to the $\varepsilon^2$ curve predicted by \cref{thm:quadratic-law} across different values of $\varepsilon$. The results are shown in \cref{fig:quadratic-rate-a}. Across the triangle, pentagon, and cube tests, the estimated log–log slopes of excess regret range from 1.980 to 2.016, close to the quadratic rate of 2 predicted by \cref{thm:quadratic-law}. As the size of the perturbations increases, the behavior of regret becomes increasingly less quadratic, as expected from the local characterization of \cref{thm:quadratic-law}. We empirically verify \cref{thm:weak}, the approximation \(\mathbb E\!\left[h(Q)^\top \mathsf K_{\lambda,s}(Q)h(Q)\right]\rightarrow\kappa_p(h)\) as $\lambda \rightarrow 0$, and at \(\lambda=2^{-9}\) they are within $5\cdot10^{-5}\%$ to $0.016\%$ of each other across the three different geometries. We provide experimental details in Appendix B.
\vspace{-0.5em}

We also test this curvature in a 24 hour battery arbitrage LP using price data from CAISO \citep{caiso_oasis_lmp_2026} and two scenario generators: a common price shift, where all prices are adjusted in the same direction, and a peak versus off-peak generator, where peak and off-peak prices vary in alternating directions. We use our scenario generation rule of \cref{cor:split} and \cref{prop:curv-gen-rule} with $\lambda=11.55$, set by a scale normalized heuristic (Appendix B). Our scenario generation rule correctly identifies the off-peak prices as most decision relevant and, as shown in \cref{fig:uncertainty-alloc-b}, allocates most of its scenario budget there. The result is a $30.8\%$ improvement in expected-cost regret compared to uniform allocation, averaged across 500 runs. 
Furthermore, at the smallest perturbation \(\varepsilon=2^{-9}\), the empirical quadratic coefficient \(2(R(f_\varepsilon)-R_X^\star)/\varepsilon^2\) differs from the analytically computed curvature \(\kappa_p(h)\) by only \(0.84\%\)–\(2.75\%\).

To conclude, these results show that the apparent nonsmoothness of LP regret has a simple local geometry at the population level. Curvature is concentrated on the decision boundaries, and can be approximated tractably with a single projection. This perspective provides both a local characterization of decision relevance and a practical mechanism for allocating computational effort toward uncertainty that can impact decisions. 
Our results also extend to (mixed) integer programs (IPs). Since IPs can be represented by an LP on its convex hull, the same quadratic behavior and curvature exist. However, estimating curvature is more difficult, since the projection and the convex hull are not readily available. An important direction is to devise an approximation that is applicable to IP problems. Beyond the scenario-allocation rule we present in this paper, other potential applications include local statistical efficiency in contextual linear optimization, decision-aware compression, and evaluation criteria for decision-focused learning. Finally, future work is to identify relaxed regularity conditions of $p(q)$ where quadratic behavior persists, quantify how the choice of reference point $s$ controls the $\mathsf K_{\lambda,s}$ approximation bias, and develop practical guidelines for selecting $\lambda$ and $s$.
\clearpage
\bibliographystyle{unsrtnat}
\bibliography{bib}
\newpage
\appendix

\section{Proofs}
\label{app:proofs}

For a set $A\subseteq\mathbb R^d$, we write $\operatorname{aff}(A)$,
$\operatorname{ri}(A)$, and $\operatorname{span}(A)$ for its affine hull\footnote{The affine hull $\operatorname{aff}(A)$ is the smallest affine subspace containing $A$, so if $y_0 \in A$, then $\operatorname{aff}(A) = y_0+ \lin(A-A)$.},
relative interior, and linear span, respectively.

\begin{lemma}[Excess regret depends only on the conditional mean]
\label{lem:conditional-mean}
Let $S$ be the (bounded) feasible set, $z_S(c)=\min_{w\in S}c^\top w$, and let
$w^\star(c)\in\arg\min_{w\in S}c^\top w$. With $Q=\mathbb{E}[C\mid X]$ and
$R_X^\star=\inf_{g:\,g(X)\in S}\mathbb{E}[C^\top g(X)-z_S(C)]$, we have
$R_X^\star=\mathbb{E}[z_S(Q)-z_S(C)]$ and, for any predictor $f$,
\[
R(f)-R_X^\star=\mathbb{E}\!\left[Q^\top w^\star(f(X))-z_S(Q)\right]\;\ge\;0 .
\]
\end{lemma}

\begin{proof}
Any decision $g(X)$ is a function of $X$ alone, so conditioning on $X$ gives
\[\mathbb{E}[C^\top g(X)]=\mathbb{E}[\mathbb{E}[C^\top g(X)|X]] = \mathbb{E}[\mathbb{E}[C |X]^\top g(X)] = \mathbb{E}[Q^\top g(X)],
\] by law of total expectations. Since $g(X)\in S$, we have $Q^\top g(X)\ge z_S(Q)$, with
equality for $g=w^\star(Q)$. Therefore
$R_X^\star=\mathbb{E}[z_S(Q)]-\mathbb{E}[z_S(C)]$. Applying the same substitution
to $g=w^\star(f(X))$ gives
$R(f)=\mathbb{E}[Q^\top w^\star(f(X))]-\mathbb{E}[z_S(C)]$, and subtracting the
two expressions cancels $\mathbb{E}[z_S(C)]$.
\end{proof}
\begin{assumption}[Local regularity]
\label{ass:local}
The relevant cost vector \(Q\) is supported on a compact set
\(K\subset\mathbb{R}^d\). For finite constants
\(L,H\), the density \(p\) and perturbation field \(h\) satisfy
\[
|p(q)-p(q')|
+
\lVert h(q)-h(q')\rVert
\le
L\lVert q-q'\rVert,
\qquad
\lVert h(q)\rVert\le H,
\]
for all \(q,q'\) in some neighborhood of $K$.
The point \(s\) lies in the relative interior of the feasible set $S$.
\end{assumption}
\begin{lemma}[Triangular crossing integral]
\label{lem:crossing}
For every $b\in\mathbb{R}$,
\[
\int_{\mathbb{R}}
|u|\,\mathbf 1\{u(u+b)<0\}\,du
=
\frac{b^2}{2}.
\]
\end{lemma}

\begin{proof}
The set $\{u:u(u+b)<0\}$ is the interval with endpoints $0$ and $-b$.
The integral is therefore the area of a right triangle with base and height
$|b|$, namely $b^2/2$.
\end{proof}

\begin{proof}[Proof of \cref{thm:quadratic-law}]
The optimizer regions of a polytope are the cones of its normal fan (see,
e.g., \citet[Ch.~7]{ziegler1995lectures}). We write $C_v$ for the cone associated with vertex $v$, i.e., the set of cost vectors for which $v$ is optimal. Two cones $C_{v_i}$ and $C_{v_j}$ share a full-dimensional common boundary, a \emph{wall} $F_E$, when $v_i, v_j$ are joined by an edge $E=[v_i,v_j]$ of $S$. We define \[
d_E := v_i - v_j, \qquad n_E := \frac{d_E}{\|d_E\|}.
\]
At a generic point of $F_E$, exactly two optimizer cones meet.
However, we first handle the degenerate case where three or more optimizer cones meet simultaneously, which happens along a lower-dimensional set $\Sigma$, of at most dimension $d-2$ and consisting of finitely many compact polyhedra.

\textbf{Step 1 (the degenerate set contributes $O(\varepsilon^3)$).}
$\Sigma$ is restricted to a fixed compact neighborhood $K$ by \cref{ass:local}. Since the normal fan is finite and polyhedral then for some $C$
\[
    \operatorname{Vol}\bigl\{q :
    \operatorname{dist}(q,\Sigma)\le C\varepsilon\bigr\}
    = O(\varepsilon^2),
\]
where it is $O(\varepsilon^2)$ since $\Sigma$ is at most codimension-two. Near a point where multiple walls meet, there are two independent directions orthogonal to $\Sigma$. Requiring $q$ to lie within $O(\varepsilon)$ in both of those directions produces the $O(\varepsilon) \times O(\varepsilon) = O(\varepsilon^2)$ volume factor.
Moreover, let
\[
    v = w^\star(q),
    \qquad
    v' = w^\star(q+\varepsilon h(q)).
\]
By optimality of $v'$ at $q+\varepsilon h(q)$,
\[
    (q+\varepsilon h(q))^\top v'
    \le
    (q+\varepsilon h(q))^\top v,
\]
and hence
\[
    0
    \le q^\top(v'-v)
    \le
    \varepsilon h(q)^\top(v-v')
    \le
    \varepsilon H\,\operatorname{diam}(S).
\]
Thus the pointwise regret is uniformly $O(\varepsilon)$. Since an
$O(\varepsilon)$-neighborhood of $\Sigma$ has volume $O(\varepsilon^2)$
and $p$ is bounded on the relevant compact set, its total contribution to
$R(f_\varepsilon)-R_X^\star$ is
\[
    O(\varepsilon)\times O(\varepsilon^2)
    =
    O(\varepsilon^3)
    =
    o(\varepsilon^2).
\]
Outside this neighborhood, any optimizer change caused by the
$O(\varepsilon)$ perturbation crosses a single wall.

\textbf{Step 2 (the rest).} It remains to treat the regular part
of each wall where exactly two optimizer cones meet. 
Let $F_E$ be a wall corresponding to the edge $E=[v_i,v_j]$, where near the regular part of $F_E$, write
\[
    q = a+t n_E,
    \qquad a\in F_E.
\]
Up to reversing $n_E$, the optimizer is $v_i$ for $t<0$ and $v_j$ for
$t>0$. If the perturbation crosses the wall, the regret is exactly
\[
\|d_E\|\,|t|,
\]
and crossing occurs precisely when
\[
t\bigl(t+\varepsilon n_E^\top h(a+t n_E)\bigr)<0.
\]
Thus the contribution of this wall is
\[
\begin{aligned}
I_{E}(\varepsilon)
&=
\int_{F_{E}}\int_{\mathbb R}
\|d_E\|\,|t|\,p(a+t n_E) \\
&\qquad\qquad\times
\mathbbm 1\!\left\{
t\bigl(t+\varepsilon n_E^\top h(a+t n_E)\bigr)<0
\right\}
\,dt\,d\mathcal H^{d-1}(a).
\end{aligned}
\]
The integrand is supported on $|t|=O(\varepsilon)$, meaning it only takes non-zero values in an area scaled by $\varepsilon$ due to the indicator term. We do a change of variables by setting $t=\varepsilon u$ and using the Lipschitz continuity of $p$ and $h$ from \cref{ass:local} which gives,
\[
\begin{aligned}
\lim_{\varepsilon\downarrow0}
\frac{I_{E}(\varepsilon)}{\varepsilon^2}
&=
\|d_E\|
\int_{F_{E}}
p(a)
\int_{\mathbb R}
|u|\,
\mathbbm 1\!\left\{
u\bigl(u+n_E^\top h(a)\bigr)<0
\right\}
du\,
d\mathcal H^{d-1}(a).
\end{aligned}
\]
Since, for every $b\in\mathbb R$ we have from \cref{lem:crossing},
\[
\int_{\mathbb R}|u|\,\mathbbm 1\{u(u+b)<0\}\,du
=
\frac{b^2}{2},
\]
we obtain
\[
\lim_{\varepsilon\downarrow0}
\frac{I_{E}(\varepsilon)}{\varepsilon^2}
=
\frac12
\int_{F_{E}}
\frac{p(a)}{\|d_E\|}
\bigl(d_E^\top h(a)\bigr)^2
\,d\mathcal H^{d-1}(a).
\]
Summing over the finitely many edges proves
\[
R(f_\varepsilon)-R_X^\star
=
\frac{\varepsilon^2}{2}
\sum_{E\in\mathcal E(S)}
\int_{F_E}
\frac{p(q)}{\|d_E\|}
\bigl(d_E^\top h(q)\bigr)^2
\,d\mathcal H^{d-1}(q)
+
o(\varepsilon^2).
\]
\end{proof}
\begin{remark}[Random perturbations used by \cref{cor:split}]
\label{rem:random}
The proof of \cref{thm:quadratic-law} also extends to random perturbations by conditioning on their additional randomness. In particular, let \(\Xi\) be independent of \(Q\), and let
\(\Delta=h(Q,\Xi)\). Assume that, for every $\xi$, the function \(q \mapsto h(q,\xi)\) is continuously differentiable near the region where $Q$ is supported, and that there exists a constant $H$ such that $\sup_\xi \sup_{q \in U}(\|h(q,\xi)\|+\|D_q h(q,\xi)\|) \leq H$ for some neighborhood $U$ of that region. In other words, both $h(q,\xi)$ and its derivative are uniformly bounded near the support of $Q$. Define the conditional second moment matrix,
\[
\Gamma(q):=\E[\Delta\Delta^\top\mid Q=q],
\]
then
\[
\E\!\left[r(Q,Q+\varepsilon\Delta)\right]
=
\frac{\varepsilon^2}{2}
\sum_{E\in\mathcal{E}(S)}
\int_{F_E}
\frac{p(q)}{\lVert d_E\rVert}
d_E^\top\Gamma(q)d_E
\,d\mathcal{H}^{d-1}(q)
+
o(\varepsilon^2).
\]
More generally, it is enough that the conditional distribution of \(\Delta\) given \(Q=q\) has the same distribution as $h(q,\Xi)$, where $\Xi$ is independent of $Q$ and $h$ satisfies the same regularity conditions. 
\end{remark}

\begin{figure}[t]
  \centering
  \begin{tikzpicture}[x=1.05cm,y=1.05cm,line join=round,line cap=round,font=\footnotesize]

  %------------------------------------------------------------------
  % cells of the normal manifold of S=[-1,1]^2, shaded by contribution:
  % edge cells (dark) carry the limit, S is O(lambda), vertex cells give 0
  %------------------------------------------------------------------
  \fill[black!18] ( 1,-1) rectangle ( 3.3, 1);
  \fill[black!18] (-1,-1) rectangle (-3.3, 1);
  \fill[black!18] (-1, 1) rectangle ( 1, 3.3);
  \fill[black!18] (-1,-1) rectangle ( 1,-3.3);

  \fill[black!5]  ( 1, 1) rectangle ( 3.3, 3.3);
  \fill[black!5]  (-1, 1) rectangle (-3.3, 3.3);
  \fill[black!5]  (-1,-1) rectangle (-3.3,-3.3);
  \fill[black!5]  ( 1,-1) rectangle ( 3.3,-3.3);

  \fill[pattern=north east lines,pattern color=black!40] (-1,-1) rectangle (1,1);

  %------------------------------------------------------------------
  % axes (thin) and walls of the manifold
  %------------------------------------------------------------------
  \draw[thin,black!55] (-3.3,0) -- (3.3,0);
  \draw[thin,black!55] (0,-3.3) -- (0,3.3);
  \node[anchor=west]  at (3.36,0) {$x_1$};
  \node[anchor=south] at (0,3.36) {$x_2$};

  \draw[line width=1.4pt] ( 1,-3.3) -- ( 1,3.3);
  \draw[line width=1.4pt] (-1,-3.3) -- (-1,3.3);
  \draw[line width=1.4pt] (-3.3, 1) -- (3.3, 1);
  \draw[line width=1.4pt] (-3.3,-1) -- (3.3,-1);
  \foreach \p in {(1,1),(-1,1),(-1,-1),(1,-1)} \fill \p circle (1.7pt);
  \node at (0,0.34) {$\bm S$};
  \node at (0,-0.34) {$\mathsf{K}_{\lambda,s}=\tfrac{1}{\lambda}I_d$};

  %------------------------------------------------------------------
  % two points in one edge cell: same affine projection
  %------------------------------------------------------------------
  \fill (0.45,2.55) circle (1.4pt);  \fill (-0.55,1.85) circle (1.4pt);
  \draw[->,>=stealth,densely dotted,black!70] (0.45,2.50) -- (0.45,1.06);
  \draw[->,>=stealth,densely dotted,black!70] (-0.55,1.80) -- (-0.55,1.06);
  \node[anchor=west,inner sep=1.5pt]  at (0.50,2.55)  {$x$};
  \node[anchor=east,inner sep=1.5pt]  at (-0.60,1.85) {$x'$};

  %------------------------------------------------------------------
  % y = Pi_S(x) on each cell, with K_{lambda,s} on one cell of each type
  %------------------------------------------------------------------
  \node at (0, 3.05)  {$y=(x_1,1)$};
  \node at (0,-2.15)  {$y=(x_1,-1)$};
  \node at (-2.15,0.34)  {$y=(-1,x_2)$};
  \node at ( 2.15,0.34) {$y=(1,x_2)$};
  \node[scale=0.85] at ( 2.15,-0.34) {$\mathsf{K}_{\lambda,s}=\tfrac{1}{\lambda}\tfrac{d_Ed_E^{\top}}{\|d_E\|^2}$};

  \node at (-2.15, 2.55) {$y=(-1,1)$};
  \node at (-2.15,-2.15) {$y=(-1,-1)$};
  \node at ( 2.15,-2.15) {$y=(1,-1)$};
  \node at ( 2.15, 2.55) {$y=(1,1)$};
  \node[scale=0.85] at ( 2.15, 2.05) {$\mathsf{K}_{\lambda,s}=0$};
  \node[scale=0.85] at (-2.15, 2.05) {$\mathsf{K}_{\lambda,s}=0$};

\end{tikzpicture}
    \caption{The normal manifold of $S=[-1,1]^2$, which partitions $\mathbb{R}^d$ into regions where the projection operator $\Pi_S$ is affine, and therefore by \cref{prop:oneproj} $\mathsf K_{\lambda,s} = \nicefrac{1}{\lambda} \Pi_{T_F}$ is constant. The points are $x=s-\nicefrac{q}{\lambda}$ and $y=\Pi_S(x)$ the resulting projection. If two points $x$ and $x'$ are in the same cell, they both project to the same face $F$ of $S$. By \cref{thm:weak}, only the edge cells (shaded dark) remain in the limit as $\lambda \downarrow 0$.}
    \label{fig:normal-manifold}
\end{figure}
\begin{proof}[Proof of \cref{thm:weak}]
For each nonempty face $F\preceq S$, write
\[
T_F:=\lin(F-F),
\qquad
k_F:=\dim F.
\]
We use the standard normal manifold description of polyhedral projection. The sets
\[
F+N_S(F), \qquad F\preceq S,
\]
form the \emph{normal manifold} of the polytope, a polyhedral subdivision
of $\mathbb R^d$ associated with Euclidean projection
\citep[Prop.~2.4.5]{scholtes2012introduction}, see also \citet{robinson1992normal} and \cref{fig:normal-manifold} for an example representation. Moreover, on the cell associated with $F$, projection onto $S$ agrees with projection onto $\operatorname{aff}(F)$ \citep[Prop.~2.4.4]{scholtes2012introduction}. Consequently, away from the lower-dimensional boundaries of the cells,
\[ J\Pi_S(x)=\Pi_{T_F},
\qquad x\in\ri F+\ri N_S(F).
\]
Since the curvature is concentrated on the walls, it is best represented by a measure, similar to the treatment of curvature for nonsmooth convex bodies in \citep{schneider1993convex}. Additionally, because $K_0$ is supported only on the fan walls, pointwise convergence is not appropriate to compare with $\mathsf K_{\lambda, s}(q)$, which is an ordinary matrix-valued function of $q$. We therefore use the standard weak formulation of measures (as in \citep[Ch.~1.9]{evans1991measure}), and we compare $\mathsf K_{\lambda,s}$ and $K_0$ through their integrals against a test function $\Phi$.

For a compactly supported Lipschitz matrix-valued function $\Phi:\mathbb{R}^d\rightarrow \mathbb{R}^{d \times d}$, set
\[I_\lambda(\Phi):=\int_{\mathbb R^d}\langle\Phi(q),\mathsf K_{\lambda,s}(q)\rangle_F\,dq,\]
where $\mathsf K_{\lambda,s}(q)$ is defined in \cref{eq:K-curvature-def}.
With
\(x=s-\nicefrac{q}{\lambda},\)
we have
\[
I_\lambda(\Phi)
=
\lambda^{d-1}
\int_{\mathbb R^d}
\left\langle
\Phi\bigl(\lambda(s-x)\bigr),
J\Pi_S(x)
\right\rangle_F\,dx,
\]
after a change of variables. 
On the normal-manifold cell of $F$, we write $x=y+n$ with
$y\in\ri F$ and $n\in\ri N_S(F)$. Since $\operatorname{span}N_S(F)=T_F^\perp$, these coordinates are orthogonal. Setting $u=-\lambda n$ therefore gives
\[
I_\lambda(\Phi)
=
\sum_{\substack{F\preceq S\\ k_F\ge1}}
\lambda^{k_F-1}
\int_{-N_S(F)}
\int_F
\left\langle
\Phi\bigl(u+\lambda(s-y)\bigr),
\Pi_{T_F}
\right\rangle_F
\,d\mathcal H^{k_F}(y)
\,d\mathcal H^{d-k_F}(u).
\]

We therefore decompose the integral in two ways: once by cell defined by the normal manifold (represented by the sum over $F$) and once by the orthogonal decomposition of $x$ into $y+n$ (represented by the two integrals, one over the cells of $F$ and one for the corresponding normal cones). 

Vertices contribute nothing to $I_\lambda(\Phi)$ because $T_F=\{0\}$, therefore $\Pi_{T_F} = 0$. For every face with
$k_F\ge2$, the prefactor is $O(\lambda)$, while compact support of
$\Phi$ keeps the remaining integral uniformly bounded. Hence these faces
contribute $O(\lambda)$, which vanishes at the limit $\lambda \rightarrow 0$.

Now let's consider the remaining case, an edge $E=[v_i,v_j]$. Writing $d_E=v_i-v_j$,
\[
\Pi_{T_E}
=
\frac{d_Ed_E^\top}{\|d_E\|^2},
\qquad
\mathcal H^1(E)=\|d_E\|,
\qquad
-N_S(E)=F_E.
\]
Since $\Phi$ is Lipschitz and $S$ is compact,
\[
\Phi\bigl(u+\lambda(s-y)\bigr)=\Phi(u)+O(\lambda)
\]
uniformly on the relevant region. Thus the contribution of $E$ is
\[
\int_{F_E}
\left\langle
\Phi(u),
\frac{d_Ed_E^\top}{\|d_E\|}
\right\rangle_F
\,d\mathcal H^{d-1}(u)
+
O(\lambda),
\]
which does not vanish as $\lambda \rightarrow 0$.
Summing over the finitely many edges gives us
\[
I_\lambda(\Phi)
=
\sum_{E\in\mathcal E(S)}
\int_{F_E}
\left\langle
\Phi(q),
\frac{d_Ed_E^\top}{\|d_E\|}
\right\rangle_F
\,d\mathcal H^{d-1}(q)
+
O(\lambda),
\]
and according to the definition in \cref{eq:K0-definition}, we see that $\mathsf K_{\lambda,s}\rightharpoonup K_0$.

Since the above is true for any compactly supported matrix-valued function, taking $\Phi(q)=p(q)h(q)h(q)^\top$, which fulfills the criteria due to \cref{ass:local}, gives
\[
\left\langle
p(q)h(q)h(q)^\top,
\frac{d_Ed_E^\top}{\|d_E\|}
\right\rangle_F
=
\frac{p(q)}{\|d_E\|}
\bigl(d_E^\top h(q)\bigr)^2,
\]
so
\[
\mathbb E\!\left[
h(Q)^\top K_{\lambda,s}(Q)h(Q)
\right]
=
\kappa_p(h)+O(\lambda).
\]
The final expansion follows from \cref{thm:quadratic-law}.
\end{proof}

\begin{proposition}[One projection gives $\mathsf K_{\lambda,s}$]
\label{prop:oneproj}
Let $y=\Pi_{S}(s-q/\lambda)$ and let $F$ be the face of $S$ whose relative interior
contains $y$. Then for almost every $q$,
\begin{equation*}
  \mathsf K_{\lambda,s}(q)=\frac{1}{\lambda}\,\Pi_{T_{F}},
  \qquad T_{F}:=\operatorname{span}(F-F),
\end{equation*}
the orthogonal projector onto the directions along $F$, scaled by $1/\lambda$.
If $S=\{w:Aw\le\beta\}$ and $A_{F}$ stacks the rows of $A$ that are active at $y$, then
$\Pi_{T_{F}}=I_{d}-A_{F}^{\top}(A_{F}A_{F}^{\top})^{\dagger}A_{F}$.
\end{proposition}

\begin{proof}[Proof of \cref{prop:oneproj}]
The projection characterization
\[
y=\Pi_S(x)
\quad\Longleftrightarrow\quad
x-y\in N_S(y)
\]
places \(x\) in the normal manifold cell
\(F+N_S(F)\). On the relative interior of this cell, the
projection is affine and tangent perturbations change \(y\) while normal perturbations do not. Therefore
\[
J\Pi_S(x)=\Pi_{\lin(F-F)},
\]
and the first formula follows from the definition of \(\Kls\) in \cref{eq:K-curvature-def}.

The row space of \(A_F\) is \(\lin(F-F)^\perp\), so
\(\ker A_F=\lin(F-F) = T_F\). The orthogonal projector onto \(\ker A_F\) is
\[
\Pi_{\ker A_F}
=
I_d-A_F^\top(A_F A_F^\top)^\dagger A_F,
\]
which proves the displayed active-set formula. See also \citep{scholtes2012introduction} and \citep{robinson1992normal}.
\end{proof}

\begin{proof}[Proof of \cref{cor:split}]
Write the error of the generated sample mean relative to the
decision-relevant conditional mean as
\[ \Delta_k:=\overline C_k-Q
= \underbrace{\mu_k-Q}_{b_k}
+ \underbrace{(\overline C_k-\mu_k)}_{\xi_k}, \]
where \(b_k:=\mu_k-Q\) is the generator bias and
\(\xi_k:=\overline C_k-\mu_k\) is the centered Monte Carlo error and \[ \overline C_k
:= \frac{1}{M_k} \sum_{m=1}^{M_k}\widetilde C_{k,m}.
\]
Conditional i.i.d.\ sampling gives
\[
\mathbb E[\xi_k\mid X]=0,
\qquad
\mathbb E[\xi_k\xi_k^\top\mid X]
=\frac{1}{M_k}\Sigma_k.
\]
Consequently,
\[
\mathbb E[\Delta_k\Delta_k^\top\mid X]
=
b_kb_k^\top+\frac{1}{M_k}\Sigma_k.
\]

Under the uniform regularity conditions of \cref{rem:random},
the random perturbation form of \cref{thm:quadratic-law}
together with \cref{thm:weak} yields
\begin{align*}
R(\overline C_k)-R_X^\star
&=
\frac12
\mathbb E\!\left[
\tr\!\left(
K_{\lambda,s}(Q)
\mathbb E[\Delta_k\Delta_k^\top\mid X]
\right)
\right]
+o(\eta_k^2)+O(\lambda\eta_k^2) \\
&=
\frac12\mathbb E\!\left[
b_k^\top K_{\lambda,s}(Q)b_k
\right]
+
\frac{1}{2M_k}
\mathbb E\!\left[
\tr\!\left(K_{\lambda,s}(Q)\Sigma_k\right)
\right]
+o(\eta_k^2)+O(\lambda\eta_k^2),
\end{align*}
which proves the claimed bias--variance decomposition. 
\end{proof}

\begin{proposition}[Curvature-weighted scenario generation rule]
\label{prop:curv-gen-rule}
For fixed contexts \(x_1,\ldots,x_n\), let
\[
a_i :=
\tr\!\left(
K_{\lambda,s}(Q_i)\Sigma_i
\right)>0.
\]
Given a continuous scenario budget \(B>0\), consider
\[
\min_{\substack{M_i>0\\ \sum_{i=1}^n M_i=B}}
\sum_{i=1}^n\frac{a_i}{M_i}.
\]
Then the unique optimal scenario generation budget allocation is
\[
M_i^\star
=
B\frac{\sqrt{a_i}}{\sum_{j=1}^n\sqrt{a_j}},
\qquad i=1,\ldots,n.
\]
\end{proposition}
\begin{proof}[Proof of \cref{prop:curv-gen-rule}]
For the generation rule, we follow the standard fixed-budget
simulation-allocation construction used in stochastic optimization and
simulation optimization (see, e.g.,
\citet{royset2013optimal} and \citet{gao2017new}).

By \cref{prop:oneproj}, \(K_{\lambda,s}\succeq 0\), so \(a_i\ge 0\).
By \cref{cor:split}, the leading average regret is, up to the common factor $1/(2n)$,
\[
\sum_{i=1}^n
\left[
b_i^\top K_{\lambda,s}(Q_i)b_i
+
\frac{a_i}{M_i}
\right].
\]
The first term is independent of the scenario counts. Hence, under the
budget constraint $\sum_{i=1}^n M_i=B$, the budget allocation problem reduces to
\[
\min_{\substack{M_i>0\\ \sum_i M_i=B}}
\sum_{i=1}^n \frac{a_i}{M_i}.
\]
By Cauchy--Schwarz,
\[
\sum_{i=1}^n\frac{a_i}{M_i}
\ge
\frac{\left(\sum_{i=1}^n\sqrt{a_i}\right)^2}{B},
\]
with equality iff \(M_i\propto\sqrt{a_i}\). Thus, whenever
\(\sum_j\sqrt{a_j}>0\),
\[
M_i^\star
=
B\frac{\sqrt{a_i}}{\sum_{j=1}^n\sqrt{a_j}}.
\]
\end{proof}
\begin{remark}[Zero curvature contexts for \cref{prop:curv-gen-rule}]
If \(a_i=0\), the scenario generation rule assigns zero budget to context \(i\). In practice, a positive minimum scenario count may be imposed in implementation, with the remaining budget allocated according to the above rule. We do so in the experiments in \cref{sec:results} and in \cref{alg:curvature-allocation} as $M_{\min}$.
\end{remark}

\section{Experimental details}
The three feasible sets, triangle $S_\Delta$, pentagon $S_5$ and cube $S_{\mathrm{cube}}$, are defined as, \[S_{\triangle}
=
\operatorname{conv}\left\{
\begin{pmatrix}-1.00\\-0.65\end{pmatrix},
\begin{pmatrix}1.00\\-0.55\end{pmatrix},
\begin{pmatrix}0.15\\1.20\end{pmatrix}
\right\},\]
\[S_{5}
=
\operatorname{conv}\left\{
\begin{pmatrix}-1.10\\-0.70\end{pmatrix},
\begin{pmatrix}0.90\\-0.80\end{pmatrix},
\begin{pmatrix}1.20\\0.25\end{pmatrix},
\begin{pmatrix}0.30\\1.15\end{pmatrix},
\begin{pmatrix}-0.95\\0.75\end{pmatrix}
\right\},\]
\[S_{\mathrm{cube}}=[-1,1]^3
=
\operatorname{conv}\left\{
(\pm1,\pm1,\pm1)
\right\}.\]
We choose the first two to be non-trivial and asymmetric so that each decision wall has different orientations, while the cube is a standard control, to test how the quadratic behavior extends to two-dimensional decision walls.
\paragraph{Computing $\kappa_p$.}
For each polytope, we compute the limiting curvature from its normal fan,
\[
\kappa_p(h)
=
\sum_{E\in\mathcal E(S)}
\int_{\mathcal F_E}
\frac{p(q)}{\|d_E\|}\bigl(d_E^\top h(q)\bigr)^2
\,d\mathcal H^{d-1}(q),
\qquad
d_E=v_i-v_j,
\]
We analytically parameterize each wall, i.e. as a ray for the 2d polygons and a 2d quadrant for each edge of the 3d cube.  The integrals are evaluated using Gauss--Legendre quadrature using the Golub-Welsch algorithm \citep{golub1969calculation,hale2013fast}, with order $128$ on each one-dimensional wall and tensor-product order $32$ for the cube. Doubling the quadrature order produced negligible changes in the reported values. We test a constant perturbation $h(q) = h_0$ and affine perturbation $h(q) = Aq + b$ and find both perform similarly. In \cref{fig:quadratic-rate-a} we show the constant perturbation for $h_0 = \nicefrac{(1,2)^\top}{\sqrt{5}}$ for the 2d feasible sets and $h_0 = \nicefrac{(1,2,3)^\top}{\sqrt{14}}$ for the 3d cube, chosen so that $\|h_0\| = 1$. To avoid a simple Gaussian density while maintaining tractability, we use the multivariate beta distribution on the unit ball for $p(q)$, where
\[p_2(q) = \frac{4}{\pi}(1-\|q\|^2)^3_+, \qquad p_3(q) = \frac{315}{64\pi}(1-\|q\|^2)^3_+,
\]
are the density functions for the 2d and 3d feasible sets respectively \citep{kabluchko2019expected,langrene2019fast}. 
\paragraph{Computing excess regret.}
For each draw $q_i\sim p$ and each of the three feasible set geometries, we compute the unperturbed and perturbed decisions
by exact enumeration of the vertices of $S$:
\[
w_i^\star\in\argmin_{w\in S}q_i^\top w,
\qquad
\widehat w_{i,\varepsilon}\in
\argmin_{w\in S}\bigl(q_i+\varepsilon h(q_i)\bigr)^\top w.
\]
The excess expected regret is estimated by
\[
\widehat{\Delta R}_{\varepsilon}
=
\frac{1}{N}\sum_{i=1}^N
q_i^\top\bigl(\widehat w_{i,\varepsilon}-w_i^\star\bigr).
\]
We use the same $q_i$ for every value of
$\varepsilon\in\{2^{-1},\ldots,2^{-9}\}$.  We use $N=2\times10^6$ samples. To assess the quadratic expansion, we compare
$2\widehat{\Delta R}_{\varepsilon}/\varepsilon^2$ with the independently
computed $\kappa_p(h)$, or equivalently plot
$2\widehat{\Delta R}_{\varepsilon}/\kappa_p(h)$ against the reference curve
$\varepsilon^2$.
\paragraph{CAISO energy arbitrage formulation.}
The energy arbitrage experiment uses hourly day ahead locational marginal prices from the CAISO OASIS system \citep{caiso_oasis_lmp_2026} for the SP15 trading hub. Each cost vector corresponds to one operating day, a 24-dimensional vector of next-day prices $\hat p_t$, which is then passed to the battery dispatch model
\begin{equation}
\label{eq:caiso-arbitrage}
g(\hat p)
\in
\arg\max_{c,d,s}
\left\{
\sum_{t=1}^{24}
\hat p_t(d_t-c_t)
-
\gamma\sum_{t=1}^{24}(c_t+d_t)
\right\}
\end{equation}
\vspace{-0.5em}
\begin{equation*}
\begin{aligned}
\text{s.t.}\quad
&s_{t+1}
=s_t+\eta_c c_t-\eta_d^{-1}d_t,
&&t=1,\ldots,24,\\
&0\le s_t\le E,
&&t=1,\ldots,24,\\
&c_t\ge 0,\quad d_t\ge 0,\quad
c_t+d_t\le P^{\max},
&&t=1,\ldots,24,\\
&s_{\mathrm{start}}=s_{\mathrm{end}}=2.
\end{aligned}
\end{equation*}
Here $c_t$, $d_t$, and $s_t$ denote charge power, discharge
power, and state of charge, respectively. We use hourly periods of unit
duration and set
\[
\gamma=5\ \text{USD/MWh},\qquad
\eta_c=\eta_d=0.95,\qquad
P^{\max}=1\ \text{MW},\qquad
E=4\ \text{MWh}.
\]
In the scenario generator experiment, for context $i$, the price vector supplied to the LP is the average of
the $M_i$ generated scenarios,
\[
\hat p_i
=
\overline p_i^{M_i}
=
\frac{1}{M_i}
\sum_{m=1}^{M_i}\widetilde p_{im}.
\]
Because the objective is linear in price, solving at this sample mean is
equivalent to optimizing the empirical expected objective. For the smoothing parameter, we use $\lambda = 11.55$, computed with the heuristic in \cref{eq:lambda-heuristic}.
\begin{equation}
\label{eq:lambda-heuristic}
\widehat{\lambda}
=
\frac{
    \frac{1}{n_G}\sum_{i=1}^{n_G}\lVert \tilde q_i\rVert
}{
    R
}
\left(
    \frac{\log \ell}{n_G}
\right)^{1/3},
\end{equation}
where $R=\sup_{w \in S} \|w-s\|$, the feasible set radius, $n_G$ the number of samples used to estimate curvature, and $\ell = \dim (\operatorname{span}(S-S))$.
We aim to provide a systematic calibration and theoretical analysis in the extended version.

For the projection, in our implementation, we eliminate state variables resulting in a battery polytope $S \subset \mathbb{R}^{48}$ in $w =(c,d)$. Prices $q \in \mathbb{R}^{24}$ are used in the minimization objective as,
\begin{equation*}
    \tilde q(q) = Gq +c_0, \qquad G = [I_{24}, -I_{24}]^\top,
\end{equation*}
where $c_0 = \gamma \mathbf{1}_{48}$. We then compute $\mathsf K_{\lambda, s}(\tilde q(q)) \in \mathbb{R}^{48\times 48}$ as per \cref{prop:oneproj} and map it back to price space as $G^\top \mathsf K_{\lambda, s}(\tilde q(q)) G$. 
\paragraph{Conditional scenario generators.}
For the CAISO battery-arbitrage experiment, each context $i$ is a 24-hour price vector $q_i$ from the CAISO sample.  Scenarios are generated from the symmetric conditional distribution
\[
\widetilde q_i=q_i+\sigma u_g\xi,
\qquad
\xi\sim\operatorname{Unif}\{-1,+1\},
\] where $q_{i,t}, t\in \{1,...,24\}$ are randomly sampled prices from a 4-year CAISO dataset, $u_g$ is a unit-norm direction and $\sigma=9.2103$ USD/MWh is set from the data as the median hourly price variability across days. The two generators then differ through the unit-norm direction. The all price-shift generator uses
\(
u_{\mathrm{level}}=\frac{\mathbf 1}{\sqrt{24}} \mathbf{1}_{24},
\)
so each scenario shifts every hourly price by approximately
$\pm1.88$ USD/MWh.  The peak-versus-off-peak generator uses
$u_{\mathrm{peak}}=v/\|v\|_2$, where
\[
v_t=
\begin{cases}
 1/5,  & t\in\{17,\ldots,21\},\\
-1/6,  & t\in\{1,\ldots,6\},\\
 0,    & \text{otherwise}.
\end{cases}
\]
Thus one sign raises peak prices by approximately $3.04$ USD/MWh while lowering off-peak prices by approximately $2.53$ USD/MWh, and the other sign reverses these changes. Both generators have covariance $\Sigma_g=\sigma^2u_gu_g^\top$, so they have exactly the same total variance, $\operatorname{tr}(\Sigma_g)=\sigma^2$.  Given an allocation of $M_i$ scenarios, the price given to the expected cost LP is their sample mean.
We generate scenarios 50 times per seed and average the regret, and repeat price sampling for 10 seeds. We allocate a generation budget of $B=10n=480$ scenarios across the two generators. 

\begin{algorithm}[t]
\caption{Curvature-weighted scenario generation}
\label{alg:curvature-allocation}
\begin{algorithmic}[1]
\Require Feasible polytope $S$, reference point $s\in\operatorname{ri}(S)$,
regularization $\lambda>0$, estimated conditional means and covariances
$\{(\widehat Q_i,\widehat\Sigma_i)\}_{i=1}^n$, total scenario budget $B$,
and allocation floor $M_{\min}$
\For{$i=1,\ldots,n$}
    \State Compute one Euclidean projection:
    \(
    w_i^\lambda
    \gets
    \Pi_S\!\left(s-\frac{\widehat Q_i}{\lambda}\right).
    \)
    \State Identify the face $F_i$ containing $w_i^\lambda$ in its
    relative interior.
    \State Compute the projector onto the directions parallel to $F_i$:
    \(
    \Pi_{T_{F_i}}
    \gets
    I-A_{F_i}^{\top}
    (A_{F_i}A_{F_i}^{\top})^\dagger A_{F_i},
    \)
    where $A_{F_i}$ contains the active constraint rows.
    \State Compute the regularized curvature:
    \(
    K_i\gets\frac{1}{\lambda}\Pi_{T_{F_i}}.
    \)
    \State Compute the curvature-weighted scenario variance:
    \(
    \alpha_i
    \gets
    \max\!\left\{
    \operatorname{tr}(K_i\widehat\Sigma_i),0
    \right\}.
    \)
\EndFor
\State Reserve the allocation floor and define the remaining budget:
\(
B_{\mathrm{rem}}\gets B-nM_{\min}.
\)
\State Compute fractional additional allocations:
\(
u_i\gets
B_{\mathrm{rem}}
\frac{\sqrt{\alpha_i}}
{\sum_{j=1}^n\sqrt{\alpha_j}},
\qquad i=1,\ldots,n.
\)
\State Initialize integer allocations:
\(
M_i\gets M_{\min}+\lfloor u_i\rfloor.
\)
\State Assign the remaining $B-\sum_iM_i$ scenarios to the contexts
with the largest fractional remainders
$u_i-\lfloor u_i\rfloor$.
\State \Return $(M_1,\ldots,M_n)$.
\end{algorithmic}
\end{algorithm}
\end{document}